\documentclass{article}
\usepackage{spconf,amsmath,graphicx}
\usepackage{multirow}
\usepackage{amssymb,amsmath,bm}
\usepackage{subfig, caption}

\title{Language Transfer of Audio Word2Vec: Learning Audio Segment Representations without Target Language Data}
\name{Chia-Hao Shen, Janet Y. Sung, Hung-Yi Lee}
\address{
National Taiwan University\\
Electrical Engineering Department\\
\{r04921047, b01901171, hungyilee\}@ntu.edu.tw
}
\begin{document}
%
\maketitle
\begin{abstract}
Audio Word2Vec offers vector representations of fixed dimensionality for variable-length audio segments using Sequence-to-sequence Autoencoder ($SA$). These vector representations are shown to describe the sequential phonetic structures of the audio segments to a good degree, with real world applications such as query-by-example Spoken Term Detection (STD). This paper examines the capability of language transfer of Audio Word2Vec. We train $SA$ from one language (source language) and use it to extract the vector representation of the audio segments of another language (target language). We found that $SA$ can still catch phonetic structure from the audio segments of the target language if the source and target languages are similar.  In query-by-example STD, we obtain the vector representations from the $SA$ learned from a large amount of source language data, and found them surpass the representations from naive encoder and $SA$ directly learned from a small amount of target language data. The result shows that it is possible to learn Audio Word2Vec model from high-resource languages and use it on low-resource languages. This further expands the usability of Audio Word2Vec. 
\end{abstract}
\section{Introduction}
\label{sec:intro}

Embedding audio word segments into fixed-length vectors has many useful applications in natural language processing such as speaker identification~\cite{IvectorIS09}, audio emotion classification~\cite{EmotionChallengeIS09}, and spoken term detection (STD)~\cite{MyJournal_SVM,segment2vectorIS13,segment2vectorIS12}. In these applications, audio segments are usually represented as feature vectors to be applied to a standard classifiers which determines the speaker's identification, emotion or whether the input queries are included. By representing the audio segments in fixed-length vectors instead of using the original segments in variable lengths, we can reduce the effort for indexing, accelerate the speed of calculation, and improve the efficiency for the retrieval task ~\cite{levin2013fixed,SRAILICASSP15,kamper2015deep}.

Recently, deep learning has been used for encoding acoustic information into vectors~\cite{WordEmbedIS14,QbyELSTMICASSP15,kamper2016deep}. 
Existing works have shown that it is possible to transform audio word segments into fixed dimensional vectors. The transformation successfully produces vector space where word audio segments with similar phonetic structures are closely located. 
In \cite{kamper2016deep}, the authors train a Siamese convolutional neural network with side information to obtain embeddings that separate same-word pairs and different-word pairs. 
Human annotated data is required under this supervised learning scenario. 
Besides supervised approaches \cite{chen2015query,kamper2016deep,he2016multi,settle2017query}, unsupervised approaches are also proposed to reduce the annotation effort \cite{chung2016audio}. 
As for the unsupervised learning for the audio embedding, LSTM-based sequence-to-sequence autoencoder  demonstrates a promising result \cite{chung2016audio}. 
The model is trained to minimize the reconstruction error of the input audio sequence and then provides the embedding, namely Audio Word2Vec, from its bottleneck layer. This is done without any annotation effort.

Although deep learning approaches have produced satisfactory result, the data-hungry nature of the deep model makes it hard to produce the same performance with low-resource data. 
Both supervised and unsupervised approaches assume that a large amount of audio data of the target language is available. 
A question arises whether it is possible to transfer the Audio Word2Vec model learned from a high-resource language into a model targeted at a low-resource language. 
While this problem is not yet to be fully examined in Audio Word2Vec, works in neural machine translation (NMT) successfully transfer the model learned on high-resource languages to low-resource languages. In \cite{zoph2016transfer, RamachandranLL16}, the authors first train a source model with high-resource language pair. The source model is used to initialize the target model which is then trained by low-resource language pairs.

For audio, all languages are uttered by human beings with a similar vocal tract structure, and therefore share some common acoustic patterns. 
This fact implies that knowledge obtained from one spoken language can be transferred onto other languages. 
This paper verifies that sequence-to-sequence autoencoder is not only able to transform audio word segments into fixed-length vectors, the model is also transferable to the languages it has never heard before. 
We also demonstrate its promising applications with a query-by-example spoken term detection (STD) experiment.
In the query-by-example STD experiment, even without tunning with partial low-resource language segments, the autoencoder can still produce high-quality vectors. 

\section{Audio Word2Vec} \label{sec:aud2vec}
The goal for Audio Word2Vec model is to identify the phonetic patterns in acoustic feature sequences such as MFCCs.
Given a sequence $\mathbf{x} = (x_{1}, x_{2}, ..., x_{T})$ where $x_{t}$ is the acoustic feature at time $t$, and $T$ is the length, Audio Word2Vec transforms the features into fixed-length vector $\mathbf{z} \in \mathbb{R}^{d}$ with dimension $d$ based on the phonetic structure. 

\subsection{RNN Encoder-Decoder Network}\label{ssec:RNN}
Recurrent Neural Networks (RNNs) has shown great success in many NLP tasks with its capability of capturing sequential information. The hidden neurons form a directed cycle and perform the same task for every element in a sequence. Given a sequence $\mathbf{x} = (x_{1}, x_{2}, ..., x_{T})$, RNN updates its hidden state $\mathbf{h}_{t}$ according to the current input $x_t$ and the previous $\mathbf{h}_{t-1}$. The hidden state $\mathbf{h}_{t}$ acts as an internal memory at time $t$ that enables the network to capture dynamic temporal information, and also allows the network to process sequences of variable length. However, in practice, RNN does not seem to learn long-term dependencies due to the vanishing gradient problem~\cite{bengio1994learning, pascanu2013difficulty}. To conquer such difficulties, LSTM~\cite{hochreiter1997long} and GRU~\cite{chung2014empirical, cho2014properties} were proposed. While LSTM achieves many amazing results ~\cite{schmidhuber2007training,Wierstra2009b,sak2014long,doetsch2014fast,cho2014learning,chung2014empirical,greff2015lstm}, the relative new GRU performs just as well with less parameters and training effort ~\cite{wu2016investigating, shang2015neural, nallapati2016abstractive, tang2016analysis}.

RNN Encoder-Decoder ~\cite{cho2014learning,sutskever2014sequence} consists of an Encoder RNN and a Decoder RNN.The Encoder RNN reads the input sequence $\mathbf{x} = (x_{1}, x_{2}, ..., x_{T})$ sequentially and the hidden state $\mathbf{h}_{t}$ of the RNN is updated accordingly. After the last symbol $x_{T}$ is processed, the hidden state $\mathbf{h}_{T}$ is interpreted as the learned representation of the whole input sequence. Then, by taking $\mathbf{h}_{T}$ as input, the Decoder RNN generates the output sequence $\mathbf{y} = (y_{1}, y_{2}, ..., y_{T'})$ sequentially, where $T$ and $T'$ can be different, or the length of $\mathbf{x}$ and $\mathbf{y}$ can be different. Such RNN Encoder-Decoder framework is able to handle variable-length input. Although there may exist a considerable time lag between the input symbols and their corresponding output symbols, LSTM and GRU are able to handle such situation well due to their powerfulness in modeling long-term dependencies.

\subsection{Sequence-to-sequence Autoencoder}    
\label{ssec:SEQ2SEQ}
\begin{figure}[h]
 \centering
 \hspace{-0.7cm}
 \includegraphics[scale=0.30]{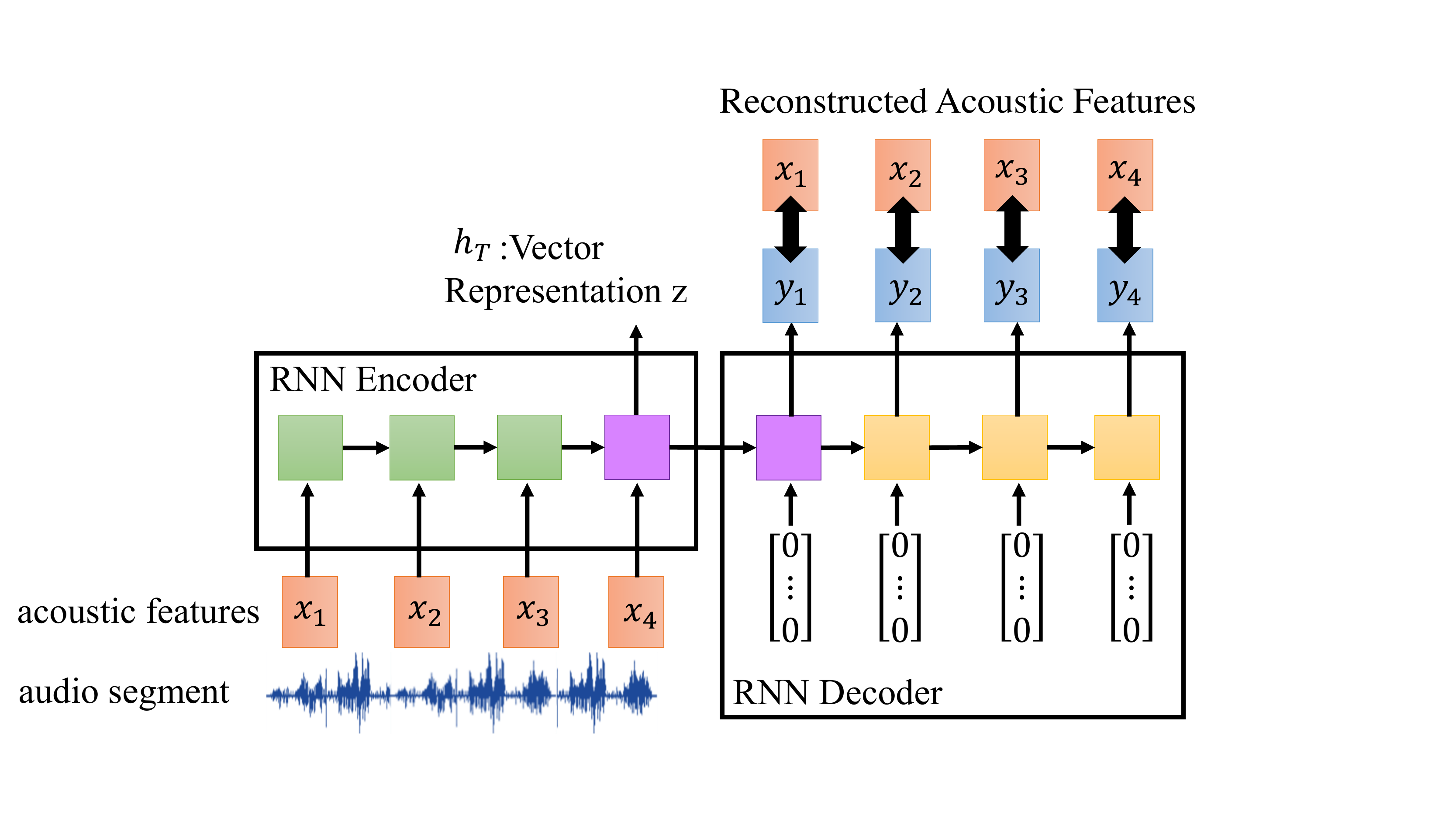}
 \caption[font=small]{Sequence-to-sequence Autoencoder ($SA$) consists of two RNNs: RNN Encoder (the left large block) and RNN Decoder (the right large block). The RNN Encoder reads an audio segment represented as an acoustic feature sequence $\mathbf{x} = (x_{1}, x_{2}, ..., x_{T})$ and maps it into a vector representation of fixed dimensionality $\mathbf{z}$; and the RNN Decoder maps the vector $\mathbf{z}$ to another sequence $\mathbf{y} = (y_{1}, y_{2}, ..., y_{T})$.
        The RNN Encoder and Decoder are jointly trained to make the output sequence $\mathbf{y}$ as close to the input sequence $\mathbf{x}$ as possible, or to minimize the reconstruction error.
}
 \label{fig:seq2seq}
 \vspace{-0.3cm}
\end{figure}

Figure~\ref{fig:seq2seq} depicts the structure of Sequence-to-sequence Autoencoder ($SA$), which integrates the RNN Encoder-Decoder framework with Autoencoder for unsupervised learning of audio segment representations. $SA$ consists of an Encoder RNN (the left part of Figure~\ref{fig:seq2seq}) and a RNN Decoder (the right part). 
Given an audio segment represented as an acoustic feature sequence $\mathbf{x} = (x_{1}, x_{2}, ..., x_{T})$ of any length $T$, the RNN Encoder reads each acoustic feature $x_{t}$ sequentially and the hidden state $\mathbf{h}_{t}$ is updated accordingly. After the last acoustic feature $x_{T}$ has been read and processed, the hidden state $\mathbf{h}_{T}$ of the Encoder RNN is viewed as the \emph{learned representation} $\mathbf{z}$ of the input sequence (the purple block in Figure~\ref{fig:seq2seq}). The Decoder RNN takes $\mathbf{h}_{T}$ as the initial state of the RNN cell, and generates a output $y_{1}$. Instead of taking $y_{1}$ as the input of the next time step, a zero vector is fed in as input to generate $y_{2}$, and so on. This structure is called the historyless decoder. Based on the principles of Autoencoder~\cite{hinton2006reducing,baldi2012autoencoders}, the target of the output sequence $\mathbf{y} = (y_{1}, y_{2}, ..., y_{T})$ is the input sequence $\mathbf{x} = (x_{1}, x_{2}, ..., x_{T})$. In other words, the RNN Encoder and Decoder are jointly trained by minimizing the reconstruction error, measured by the general mean squared error $\sum_{t=1}^{T}\| x_{t} - y_{t} \|^2$. Because the input sequence is taken as the learning target, the training process does not need any labeled data. The fixed-length vector representation $\mathbf{z}$ will be a meaningful representation for the input audio segment $\mathbf{x}$ because the whole input sequence $\mathbf{x}$ can be reconstructed from $\mathbf{z}$ by the RNN Decoder.

Using historyless decoder is critical here. 
We found out that the performance in the STD experiment was undermined despite the low reconstruction error. 
This shows that the vector representations learned from $SA$ do not include useful information.
This might be caused by a strong decoder as the model focuses less on including more information into the vector representation. We eventually solved the problem by using a historyless decoder.
Historyless decoder is a weakened decoder. 
The input of the decoder is removed, and this forces the model to rely more on the vector representation. 
The historyless decoder is also used in recent NLP works \cite{bowman2016generating, semeniuta2017hybrid, nema2017diversity}.

\section{Language Transfer}
\label{sec:ltransfer}
\begin{figure}[h]
 \centering
 \hspace{-0.5cm}  \includegraphics[scale=0.265]{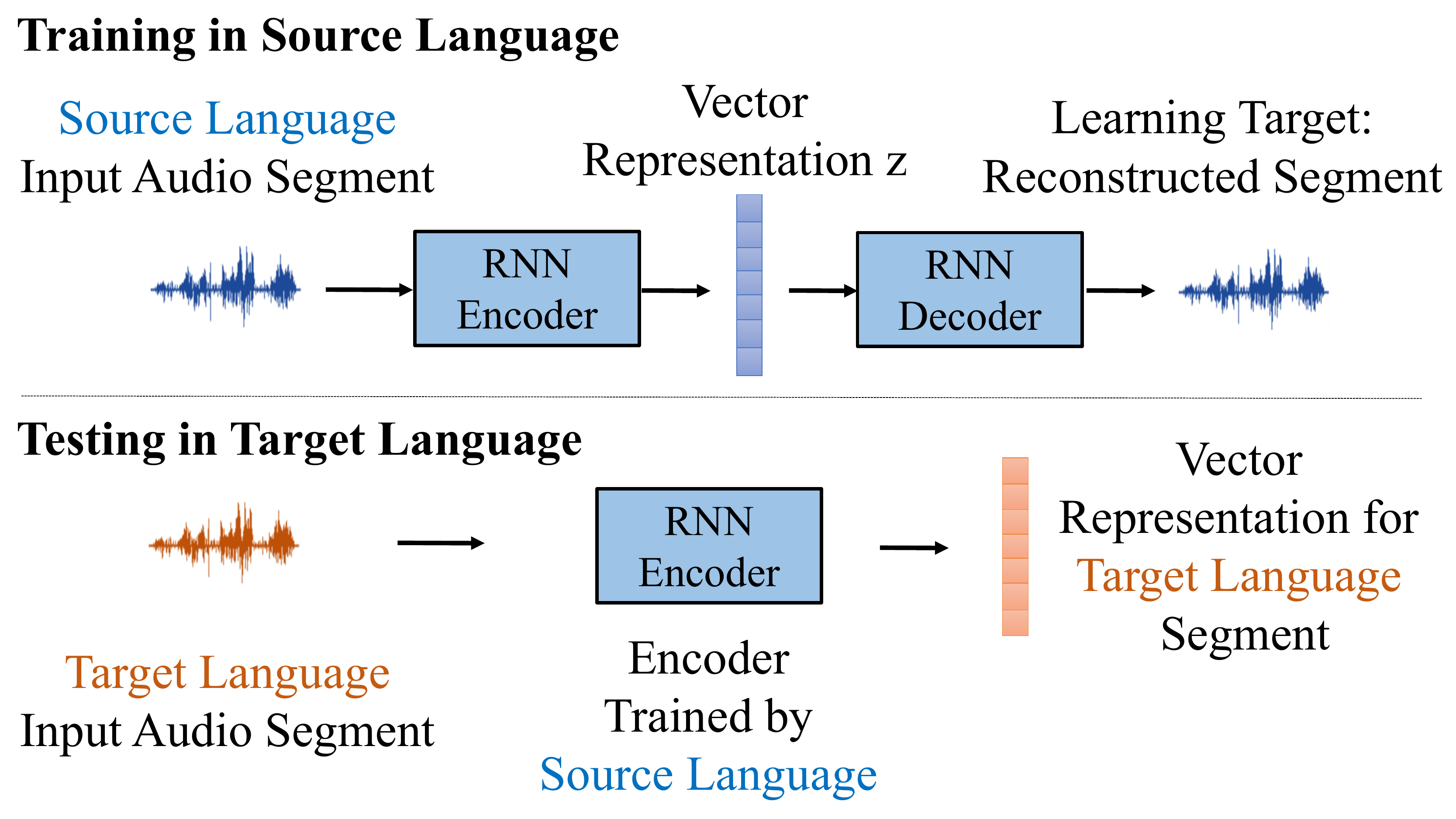}
 \vspace{-0.6cm}
 \caption[font=small]{
 The training and testing mechanism for language transfer. In the training phase, a RNN Encoder-Decoder is trained by an abundant amount of audio segments in the source language, which is shown in the upper section of the figure and marked blue. During the testing phase, the learned RNN Encoder for the source language (the blue block in the lower section of the figure) is directly used to transform an audio segment in the target language (the orange segment) to a fixed-length vector.
 }
 \label{fig:transfer}
\end{figure}

In the study of linguistic, scholars define a set of universal phonetic rules which describe how sounds are commonly organized across different languages. Actually, in real life, we often find languages sharing similar phonemes especially the ones spoken in nearby regions. These facts implies that when switching target languages, we do not need to learn the new audio pattern from scratch due to the transferability in spoken languages. 
Language transfer has shown to be helpful in STD~\cite{GTTSMediaEval13,GTTSMediaEval14,MultipleTokenICASSP13,NNIMediaEval14,CUHKMediaEval12,CUHKMediaEval14,SpeeDMediaEval2014,SPLMediaEval14}.
In this paper, we focus on studying the capability of transfer learning of Audio Word2Vec.

In the proposed approach, we first train an $SA$  using the high-resource source language, as shown in the upper part of Fig.~\ref{fig:transfer}, and then the encoder is used to transform the audio segment of a low-resource target language. 
It is also possible to fine-tune the parameters of $SA$ with the target language. 
In the following experiments, we found that in some cases the STD performance of the encoder without fine-tuning with the low-resource target language can be as good as the one with fine-tuning. 

\section{An Example Application: Query-by-example STD}
\label{sec:STD}

  \begin{figure}[h]
    \vspace{-0.5cm}
    \includegraphics[scale=0.25]{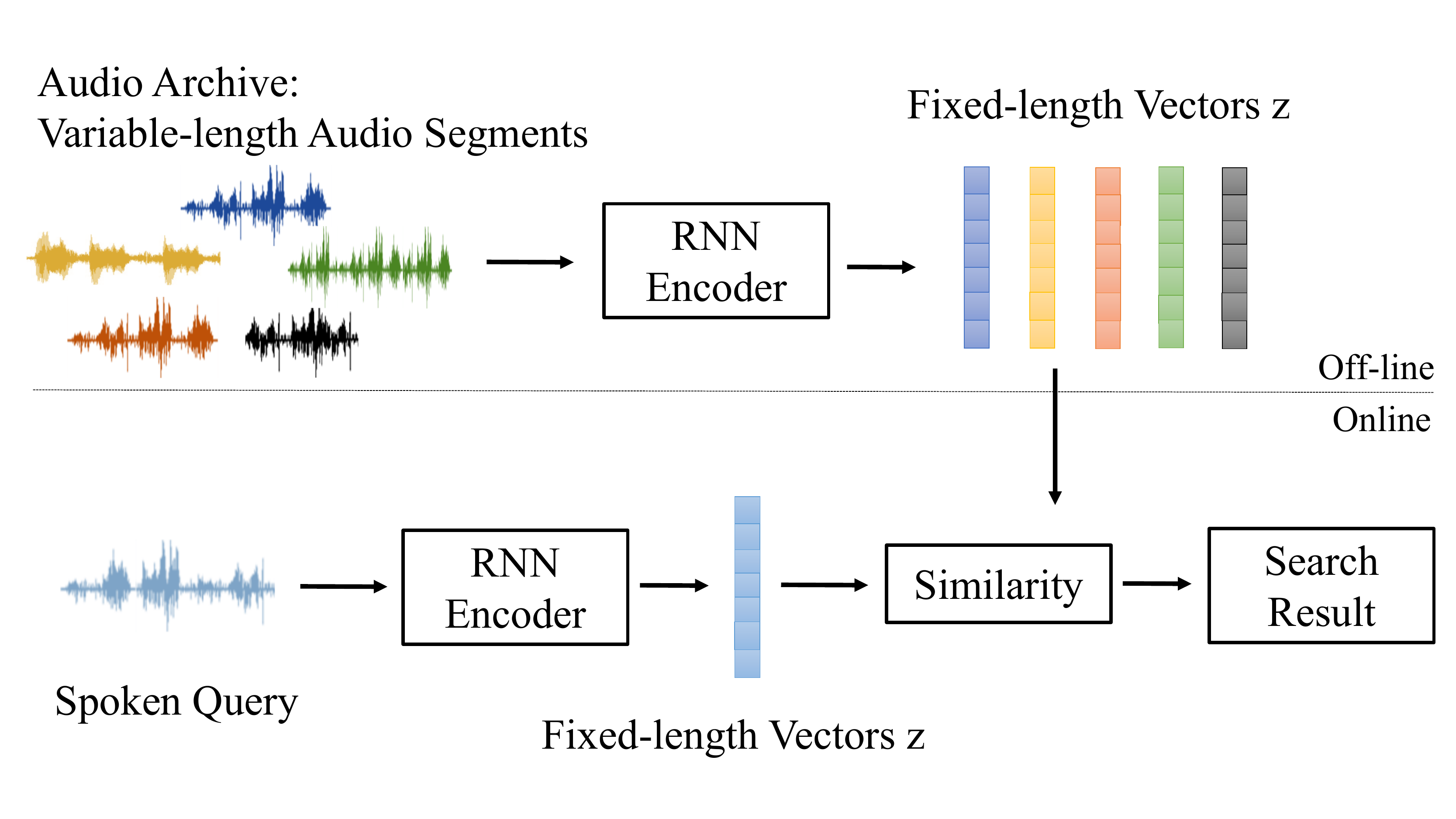}
    \vspace{-0.5cm}
    \caption{
          The example application of query-by-example STD.
      All audio segments in the audio archive are segmented based on word boundaries and represented by fixed-length vectors off-line.
      When a spoken query is entered, it is also represented as a vector. The similarity between this vector and all vectors for segments in the archive are calculated, and the audio segments are ranked accordingly.
    }
    \label{fig:qbe}
  \end{figure}

  The audio segment representation $\mathbf{z}$ learned in the last section can be applied in many possible scenarios. Here in the preliminary tests we consider the unsupervised query-by-example STD, whose target is to locate the occurrence regions of the input spoken query term in a large spoken archive without speech recognition. Figure~\ref{fig:qbe} shows how the representation $\mathbf{z}$ proposed here can be easily used in this task. This approach is inspired from the previous work~\cite{SRAILICASSP15}, but completely different in the ways to represent the audio segments. In the upper half of Figure~\ref{fig:qbe}, the audio archive are segmented based on word boundaries into variable-length sequences, and then the system exploits the trained RNN encoder in Figure~\ref{fig:seq2seq} to encode these audio segments into fixed-length vectors. All these are done off-line. In the lower left corner of Figure~\ref{fig:qbe}, when a spoken query is entered, the input spoken query is similarly encoded by the same RNN encoder into a vector. The system then returns a list of audio segments in the archive ranked according to the cosine similarities evaluated between the vector representation of the query and those of all segments in the archive. Note that the computation requirements for the online process here are extremely low.

\section{Experimental Setup}
\label{sec:exp_setup}
Here we provide detail of our experiment including the dataset, model setup, and the baseline model.
\subsection{Dataset}
\label{ssec:exp_data}

Two corpora across five languages were used in the experiment. 
One of the corpora we used is LibriSpeech corpus \cite{panayotov2015librispeech} (English). 
In this 960-hour English dataset, 2.2 million audio word segments were used for training while the other 250 thousand segments were used as the database to be retrieved in STD and 1 thousand segments as spoken queries. 
In Section 6.1, we further sampled 20 thousand segments from 250 thousand segments to form a small database to investigate the influence of database size. 
English served as the high-resource source language for model pre-training.

The other dataset is the GlobalPhone corpus \cite{schultz2002globalphone}, which includes French (FRE), German (GER), Czech (CZE), and Spanish (ESP). 
The four languages from GlobalPhone were used as the low-resource target languages.
In Section 6.2, 20 thousand segments for each language were used to calculate the average cosine similarity.
For the experiments of STD, the 20 thousands segments served as the database to be retrieved, and the other 1 thousand used for query and 4 thousand for fine-tuning.



MFCCs of 39-dim were used as the acoustic features. 
The length of the  input sequence was limited to 50 frames. 
All datasets were segmented according to the word boundaries obtained by forced alignment with respect to the reference transcriptions.
Although the oracle word boundaries were used here for the query-by-example STD in the preliminary tests, the comparison in the following experiment was fair since all approaches used the same segmentation.
Mean average precision (MAP) was used as the evaluation measure for query-by-example STD.

\subsection{Proposed Model: Sequence Autoencoder ($SA$)}
Both the proposed model ($SA$) and baseline model ($NE$, described in the next subsection) were implemented with Tensorflow. The network structure and the hyper parameters were set as below:

\begin{itemize}
\item Both RNN Encoder and Decoder consisted one hidden layer of GRU cells \cite{chung2014empirical, cho2014properties}. The number of units in the layer would be discussed in the experiment.
\item The networks were trained by SGD without momentum. The initial learning rate was 1 and decayed with a factor of 0.95 every 500 batches.
\end{itemize}

\subsection{Baseline: Naive Encoder ($NE$)}
\label{ssec:baseline}
We used naive encoder ($NE$) as the baseline approach.
In this encoder, the input acoustic feature sequence $\mathbf{x}$ = ($x_1, x_2, x_3, ..., x_T$), where $x_t$ was the 39-dimension MFCC feature vector at time t, were divided into $m$ partitions with roughly equal length $T/m$. Then, we averaged each partition into a single 39-dimension vector, and finally got the vector representation through concatenating the $m$ average vectors sequentially into a vector representation of dimensionality $ 39 \times m $. 
Although $NE$ is simple, similar approaches have been used in STD and achieved successful results~\cite{MyJournal_SVM,segment2vectorIS13,segment2vectorIS12}. 

\section{Experiments}
\label{sec:experiments}
In this section, we first examine how changing the hidden layer size of the RNN Encoder/Decoder, the dimension of Audio Word2Vec, affects the MAP performance of query-by-example STD (Section 6.1). After obtaining the best hidden layer size, we analyze the transferability of the Audio Word2Vec by comparing the cosine similarity of the learned representations to phoneme sequence edit distance (Section 6.2) . Visualization of multiple word pairs in different target languages is also provided (Section 6.3). Last but not least, we performed the query-by-example STD on target languages (Section 6.4). These experiments together verify that $SA$ is capable of extracting common phonetic structure in human language and thus is transferable to various languages.

\subsection{Analysis on Dimension of Audio Word2Vector}\label{ssec:STD}

\begin{figure}[h]
 \hspace{-0.3cm}  \includegraphics[scale=0.55]{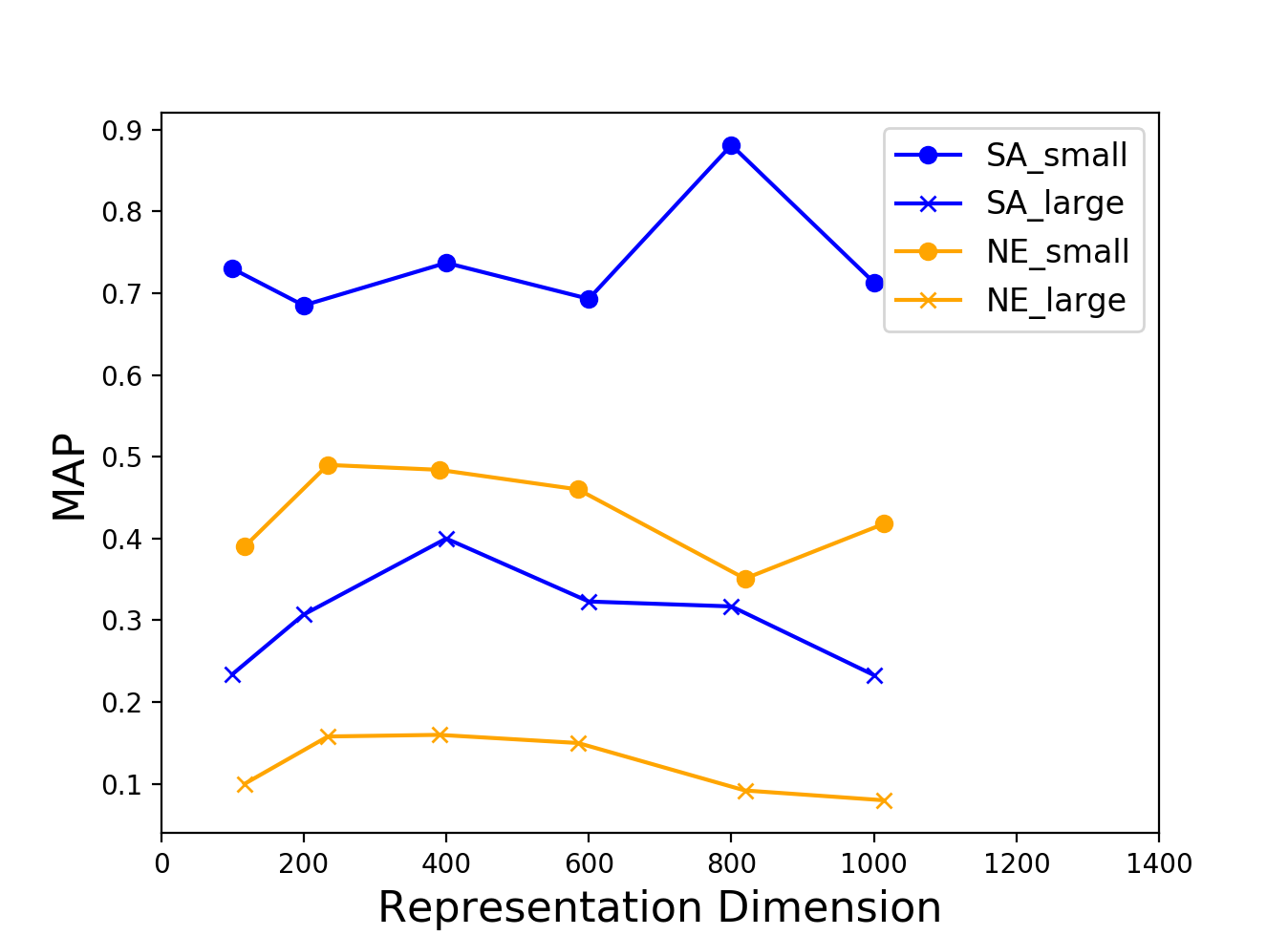}
 \caption[font=small]{
 The retrieval performance in MAP for $NE$ and $SA$ with different dimensions on large testing database (250K segments) and small database (20K).
 }
 \label{fig:source_map}
\end{figure}

Before evaluating the language transfer result, we first experimented on the primary $SA$ model in the source language (English).
The results are shown in Fig.~\ref{fig:source_map}.
Here we compare the representations of $SA$ and $NE$.
Furthermore, we examined the influence of the dimension of Audio Word2Vector in terms of MAP.
We also compared the MAP results on large testing database (250K segments) and small database (20K).

In Fig.~\ref{fig:source_map}, we varied the dimension of Audio Word2Vector as 100, 200, 400, 600, 800 and 1000. 
To match up the dimensionality with $SA$, we tested $NE$ with dimensionality 117, 234, 390, 585, 819, 1014   ($m=3,6,10,15,21,26$) and denoted them by $NE_{d}$ where $d$ is the dimensionality. 
$SA$ get higher MAP values than $NE$ no matter the vector dimension and the size of database.
The highest MAP score $SA$ can achieve is 0.881 ($SA_{800}$ on small database), while the highest score of the $NE$ model is 0.490 ($NE_{234}$ on small database).
The size of database has large influence on the results.
The MAP scores of the two models both drop in the large database. For example, $NE_{234}$ drops from 0.490 to 0.158, decaying by 68\%, and the performance of $SA_{800}$ drops from 0.881 to 0.317, decaying by 64\%. 
As shown in Fig. \ref{fig:source_map}, larger dimensionality does not imply better performance in query-by-example STD. The MAP scores gradually improve until reaching the dimensionality of 400 in $SA$ and 234 in $NE$, and start to decrease as the dimension increases. 
In the rest of the experiments, we would use 400 GRU units in the $SA$ hidden layer, and set $NE=NE_{234}$ ($m=6$).

\subsection{Analysis of Language Transfer}\label{ssec:representation}
\begin{figure}[h]
 \hspace{-0.2cm}  \includegraphics[scale=0.5]{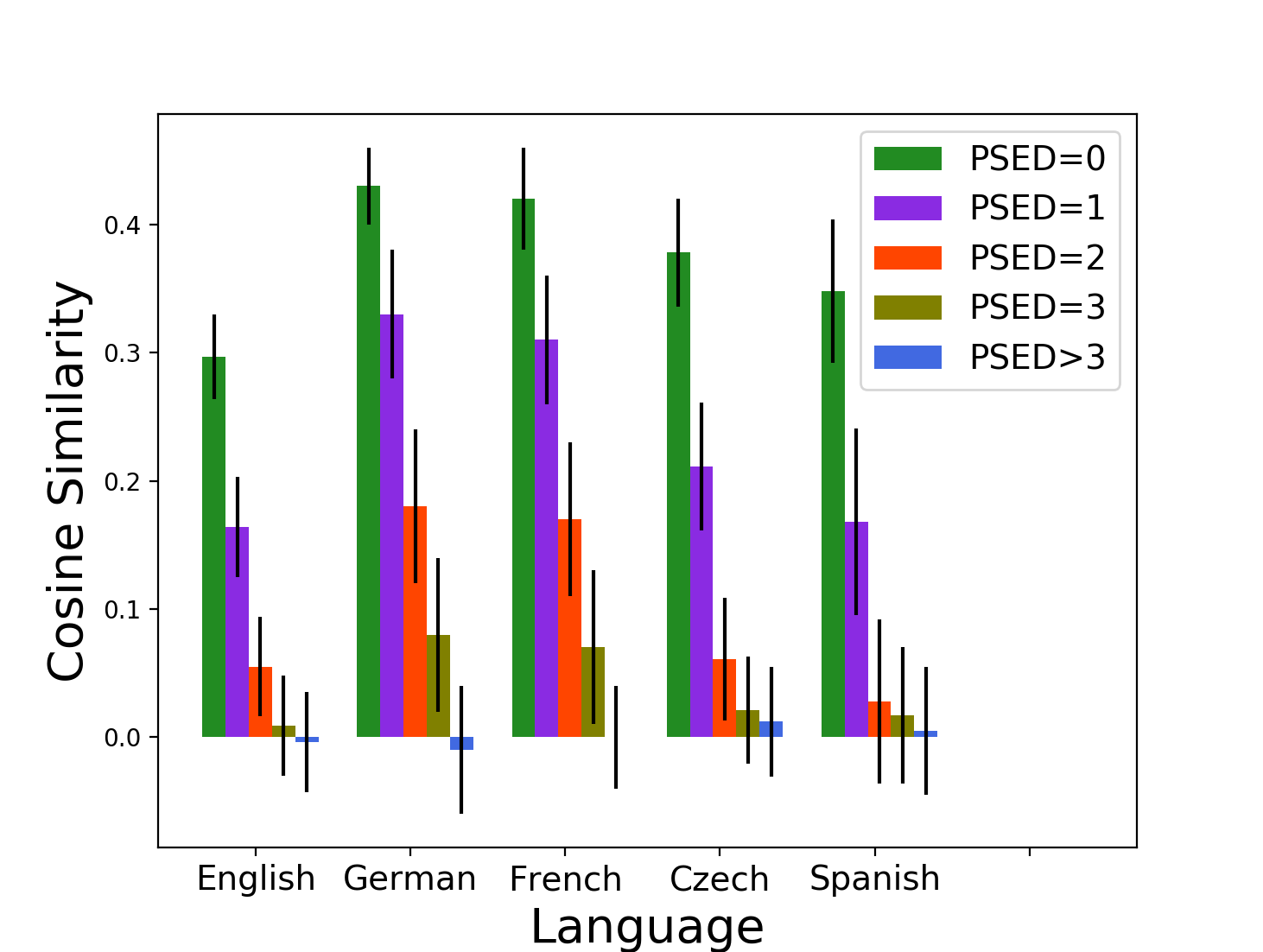}
 \vspace{-0.45cm}
 \caption[font=small]{
 The average cosine similarity and variance (the length of the black line on each bar) between the vector representations for all the segment pairs in the target languages testing set, clustered by the phoneme sequence edit distances (PSED).
 }
 \label{fig:cosine_sim}
 \vspace{-0.3cm}
\end{figure}

To evaluate the quality of language transfer, we trained the Audio Word2Vec model by $SA$ from the source language, English, and applied it on different target languages, French (FRE), German (GER), Czech (CZE), and Spanish (ESP).
We computed the average cosine similarity of the vector representations for each pair of the audio segments in the retrieval database  of the target languages (20K segments for each language), and compare it with the phoneme sequence edit distance (PSED). 
The average and variance (the length of the black line on each bar) of the cosine similarity for groups of pairs clustered by the phoneme sequence edit distances (PSED) between the two words are shown in Fig. \ref{fig:cosine_sim}.
For comparison, we also provide the results obtained from the English retrieval database (250K segments), where the segments were not seen by the model in training procedure. 

In Fig. \ref{fig:cosine_sim}, the cosine similarities of the segment pairs get smaller as the edit distances increase, and the trend is observed in all languages. The gap between each edit distance groups, i.e. (0,1), (1,2), (2,3), (3,4), is obvious. 
This means that $SA$ learned from English can successfully encode the sequential phonetic structures into fixed-length vector for the target languages to some good extend even though \textit{it has never seen any audio data of the target languages}.  

Another interesting fact is the corresponding variance between languages. 
In the source language, English, the variances of the five edit distance groups are fixed at 0.030, which means that the cosine similarity in each edit distance group is centralized. 
However, the variances of the groups in the target languages vary. 
In French and German, the variance grows from 0.030 to 0.060 as the edit distance increases from 0 to 4. 
For Czech/Spanish, the variance starts at a larger value of 0.040/0.050 and increases to 0.050/0.073.
We suspect that the fluctuating variance is related to the similarity between languages.
English, German and French are more similar compared with Czech and Spanish. 
Among the four target languages, German has the highest lexical similarity with English (0.60) and the second highest is French (0.27),  while for Czech and Spanish, the lexical similarity scores is 0 \cite{ethnologue}.

\vspace{-0.4cm}

\subsection{Visualization}
\label{ssec:visualize}

In order to further investigate the performance of $SA$, we visualize the vector representation of two sets of word pairs differing by only one phoneme from French and German as below: 

\begin{enumerate}
\item French Word Pairs: (parler, parlons), (noter,notons), (rappeler, rappelons), (utiliser, utilisons)
\item German Word Pairs: (tag, tage), (spiel, spiele), (wenig, wenige), (angriff, angriffe)
\end{enumerate}

\begin{figure}[t]
 \centering
 \subfloat[French word pairs: the last phoneme changes from `er' to `ons'. \label{fig:fre_vis}]{\includegraphics[scale=0.3]{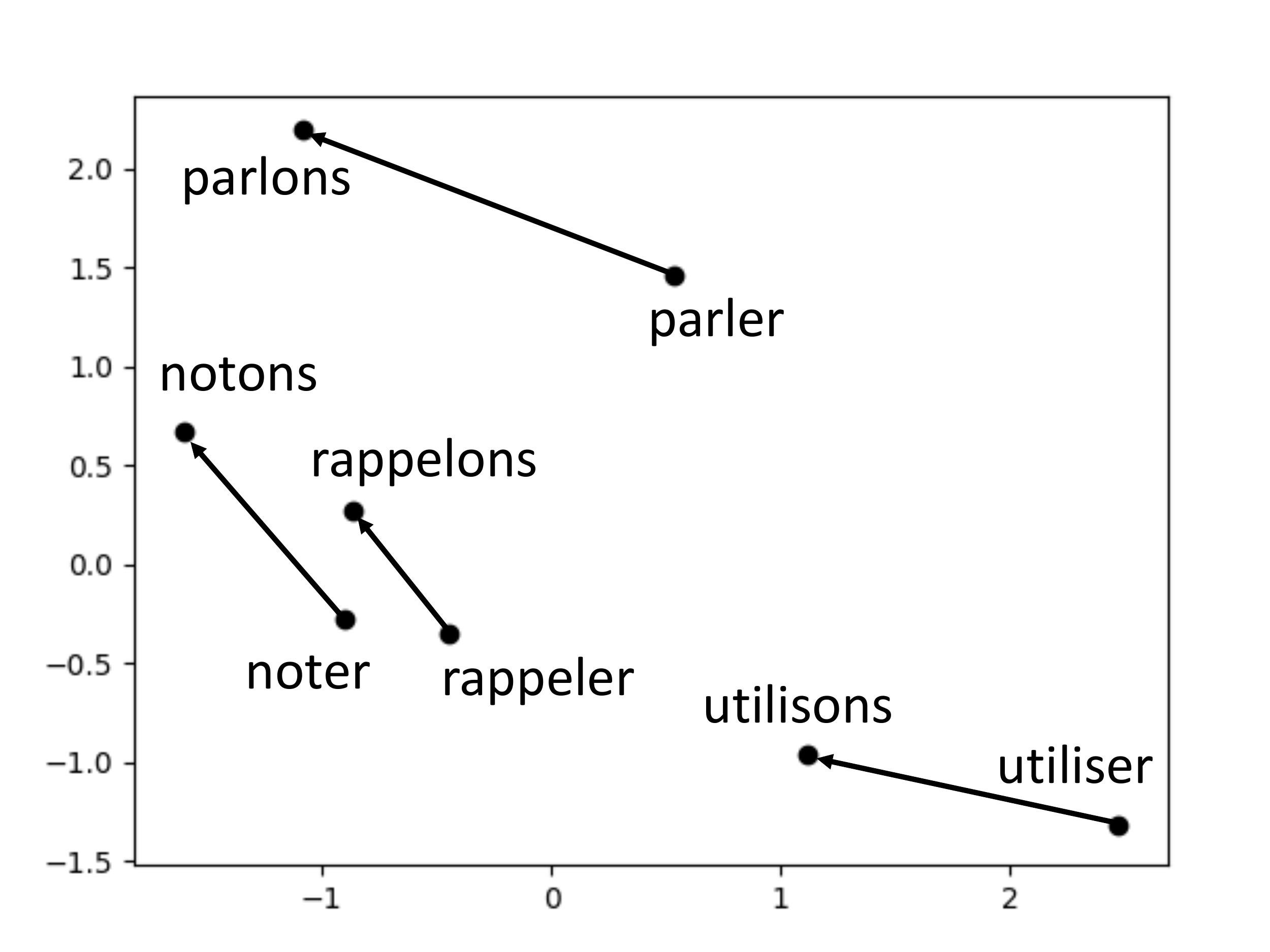}} \hfill
 \subfloat[German word pairs: the last phoneme differs by existing `e' or not. \label{fig:ger_vis}]{\includegraphics[scale=0.3]{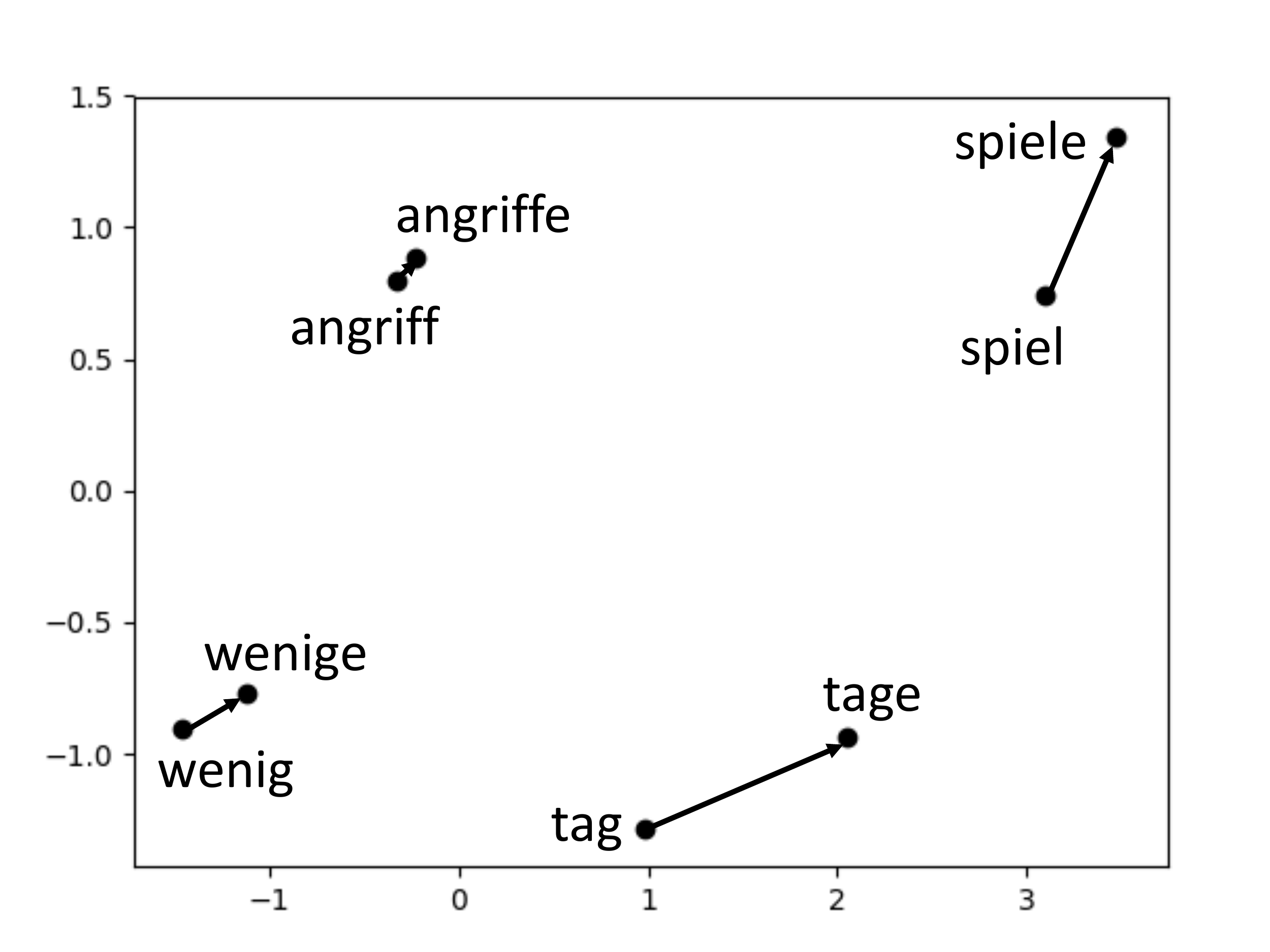}}
 \hspace{0cm}
 \caption[font=small]{Difference vectors between the average vector representations for word pairs differing by one edit distance in (a) French and (b) German.}
 \label{fig:vis}
 \vspace{-0.3cm}
\end{figure}

To show the vector representations in Fig.~\ref{fig:vis}, we first obtained the mean value of representations for the audio segments of a specific word, denoted by $\delta$(word). Then the average representation $\delta$ was projected from 400-dimensional to 2-dimensional using PCA \cite{mikolov2013distributed}. The result of the difference vector from each word pair, e.g. $\delta$(parlons) - $\delta$(parler), is shown. 
Although the representations for French and German word audio segments were extracted from the model trained by English audio word segments and never heard any French and German, the direction and magnitude of the different vectors are coherent. 
In Fig.~\ref{fig:fre_vis}, $\delta$(parlons) - $\delta$(parler) is close to $\delta$(utilison) - $\delta$(utiliser); and $\delta$(tage) - $\delta$(tag) is close to $\delta$(wenige) - $\delta$(wenig) in Fig.~\ref{fig:ger_vis}. 
  
\subsection{Language Transferring on STD} \label{ssec:transfer}
Besides analyzing the cosine similarity of the learned representations, we also apply them to the query-by-example STD task. 
Here we compare the retrieval performance in MAP of $SA$ with different levels of accessibility to the low-resource target language along with two baseline models, $NE$ and $SA$ trained purely by the target languages. 
For the four target languages, the total available amount of audio word segments in the training set were 4 thousands for each language. In Table~\ref{tab:STD_trans}, we took different partitions of the target language training sets to fine tune the $SA$ pretrained by the source languages. The amount of audio word segments in these partitions are: 1K, 2K, 3K, 4K, and 0, which means no fine-tuning.

From Table~\ref{tab:STD_trans}, $SA$ trained by source language generally outperforms the $SA$ trained by the limited amount of target language ("$SA$ No Transfer"), proving that with enough audio segments, $SA$ can identify and encode universal phonetic structure. 
Comparing with NE, $SA$ surpasses $NE$ in German and French even without fine-tuning, whereas in Czech, $SA$ also achieves better score than $NE$ with fine-tuning. 
However, in Spanish, $SA$ achieved a MAP score of 0.13 with fine-tuning, slightly lower than 0.17 obtained by $NE$. 
Back to Fig.~\ref{fig:cosine_sim}, the gap between phoneme sequence edit distances 2 and 3 in Spanish is smaller than  other languages. 
Also, as discussed earlier in Section 6.2, the variance in Spanish is also bigger. 
The smaller gap and bigger variance together indicate that the model is weaker on Spanish  at identifying audio segments of different words and thus affects the MAP performance in Spanish.

\begin{table}[h!]
 \begin{center}
    \caption{
    The retrieval performance of $NE$, $SA$ trained by the target language only (denoted as $SA$ No Transfer), and $SA$ of the source language tuning with different amounts of data. The numbers (0, 1K, 2K, 3K, 4K) are the amount of target language segments used to tune the original $SA$ trained by the source language. For example, $SA$ 2K means that the $SA$ is first trained by the source language and then tuned by 2K target language segments.  
    }
    \label{tab:STD_trans}
    \begin{tabular}{|c|c|c|c|c|c|}
    \cline{3-6}
    \multicolumn{2}{c|}{ } & FRE & GER & CZE & ESP \\
    \hline\hline
    \multicolumn{2}{|c|}{ NE } & 0.22 & 0.18 & 0.09 & \textbf{0.17}\\
    \hline\hline
    \multicolumn{2}{|c|}{ \begin{tabular}[c]{@{}c@{}}SA\\ No Transfer \end{tabular} } & 0.03 & 0.01 & 0.00 & 0.00 \\
    \hline\hline
    \multirow{5}{*}{SA}
    & 0 & \textbf{0.26} & 0.24 & 0.06  & 0.04\\
    \cline{2-6}
    & 1K & 0.24 & 0.20 & 0.09 & 0.13 \\ 
    \cline{2-6}
    & 2K & \textbf{0.26} & \textbf{0.25} & \textbf{0.10} & 0.12  \\ 
    \cline{2-6}
    & 3K & 0.22 & 0.19 & 0.08 & 0.11\\
    \cline{2-6}
    & 4K & \textbf{0.26} & 0.20 & 0.09 & 0.13\\
    \hline
    \end{tabular}
 \end{center}
\end{table}

\section{Conclusion and Future Work}
\label{sec:conclusion}
In this paper, we verify the capability of language transfer of Audio Word2Vec using Sequence-to-sequence Autoencoer ($SA$). 
We demonstrate that $SA$  can learn the sequential phonetic structure commonly appearing in human language and thus make it possible to apply an Audio Word2Vec model learned from high-resource language to low-resource languages. 
The capability of language transfer in Audio Word2Vec is beneficial to many real world applications, for example, the query-by-example STD shown in this work. For the future work, we are examining the performance of the transferred system in other application scenarios, and exploring the performance of Audio Word2Vec under automatic segmentation.


\bibliographystyle{IEEEbib}
\bibliography{strings,refs,interspeech-16,IR_bib}

\end{document}